\documentclass[letterpaper, 10 pt, conference]{ieeeconf}
\IEEEoverridecommandlockouts    
\usepackage{graphics} 
\usepackage{epsfig} 
\usepackage{times} 
\usepackage{amsmath} 
\usepackage{amssymb}  

\usepackage{mathtools}
\usepackage{color}
\usepackage{cite}
\usepackage{comment}
\usepackage{balance}
\usepackage[table]{xcolor}
\usepackage{multirow}
\usepackage{hyperref}
\hypersetup{
    colorlinks=true,
    linkcolor=black,
    citecolor=green,
    urlcolor=blue,
}
\usepackage{blindtext}
\usepackage{bm}
\usepackage{algorithm}
\usepackage{algpseudocode}
\usepackage[official]{eurosym}
\usepackage{subcaption}
\usepackage{booktabs}
\usepackage{makecell}  

\algrenewcommand\algorithmicindent{1.0em}
\algnewcommand\algorithmicswitch{\textbf{switch}}
\algnewcommand\algorithmiccase{\textbf{case}}
\algnewcommand\algorithmicassert{\texttt{assert}}
\algnewcommand\Assert[1]{\State \algorithmicassert(#1)}%
\algdef{SE}[SWITCH]{Switch}{EndSwitch}[1]{\algorithmicswitch\ #1\ \algorithmicdo}{\algorithmicend\ \algorithmicswitch}%
\algdef{SE}[CASE]{Case}{EndCase}[1]{\algorithmiccase\ #1}{\algorithmicend\ \algorithmiccase}%
\algtext*{EndSwitch}%
\algtext*{EndCase}%

\makeatletter
\newcommand{\algmargin}{\the\ALG@thistlm}
\makeatother
\newlength{\forwidth}
\algdef{SE}[parFOR]{parFor}{EndparFor}[1]
  {\parbox[t]{\dimexpr\linewidth-\algmargin}{%
     \hangindent\forwidth\strut\algorithmicfor\ #1\ \algorithmicdo\strut}}{\algorithmicend\ \algorithmicfor}%
\newlength{\forif}
\algnewcommand{\parState}[1]{\State%
  \parbox[t]{\dimexpr\linewidth-\algmargin}{\strut #1\strut}}

\title{\LARGE \bf
Pedestrian Motion Prediction Using Transformer-based \\ Behavior Clustering and Data-Driven Reachability Analysis
}

\author{Kleio Fragkedaki, Frank J. Jiang, Karl H. Johansson, Jonas Mårtensson%
\thanks{
    This work was partially supported by the Swedish Innovation agency (Vinnova), under grant 2021-02555 Future 5G Ride, within the Strategic Vehicle Research and Innovation program (FFI); the Wallenberg Artificial Intelligence, Autonomous Systems, and Software Program (WASP) funded by the Knut and Alice Wallenberg Foundation; the Swedish Research Council Distinguished Professor Grant 2017-01078; and the Knut and Alice Wallenberg Foundation Wallenberg Scholar Grant.}
\thanks{
    All authors are with the Division of Decision and Control Systems, EECS, KTH Royal Institute of Technology, Malvinas v{\"a}g 10, 10044 Stockholm, Sweden {\tt\small \{kfrag, frankji, kallej, jonas1\}@kth.se}. They are also affiliated with the Integrated Transport Research Lab and Digital Futures.}%
}

\begin{document}

\maketitle
\thispagestyle{empty}
\pagestyle{empty}

\begin{abstract}
In this work, we present a transformer-based framework for predicting future pedestrian states based on clustered historical trajectory data. In previous studies, researchers propose enhancing pedestrian trajectory predictions by using manually crafted labels to categorize pedestrian behaviors and intentions. However, these approaches often only capture a limited range of pedestrian behaviors and introduce human bias into the predictions. To alleviate the dependency on manually crafted labels, we utilize a transformer encoder coupled with hierarchical density-based clustering to automatically identify diverse behavior patterns, and use these clusters in data-driven reachability analysis. By using a transformer-based approach, we seek to enhance the representation of pedestrian trajectories and uncover characteristics or features that are subsequently used to group trajectories into different ``behavior'' clusters. We show that these behavior clusters can be used with data-driven reachability analysis, yielding an end-to-end data-driven approach to predicting the future motion of pedestrians. We train and evaluate our approach on a real pedestrian dataset, showcasing its effectiveness in forecasting pedestrian movements. 
\end{abstract}

\section{Introduction}\label{sec:intro}

Today, pedestrians continue to pose a challenge for ensuring the safety of intelligent transport systems. Pedestrians are able to freely move around in the same space as vehicles and often cross paths with vehicles. In such scenarios, the vehicles need to ensure the safety of the pedestrians, regardless of the pedestrians' actions. This poses a significant challenge for automated vehicles since pedestrians are both difficult to model and have a large degree of freedom. Due to this challenge, ensuring their safety often results in overly conservative driving policies.

A popular approach for enhancing the safety of intelligent transport systems is the integration of reachability analysis~\cite{Althoff2014, Pek2020, Söderlund2023}. By using reachability analysis, we are able to predict an over-approximation of the set of all possible places a pedestrian can be in the future, regardless of what specific decisions they make. Then, we can incorporate these sets into the path planning and control system of a vehicle, so it can make safe decisions around pedestrians. However, when reachability analysis is applied to predict the future motion of pedestrians, two issues arise: (1) there is little consensus on accurate models for pedestrian motion and (2) even if a model is chosen, the resultant reachable sets end up being very conservative due to the pedestrians' decision freedom. To address the first issue, several have proposed the use of data-driven reachability analysis for agents that are difficult to model~\cite{Devonport2021, Alanwar2023}. The use of data-driven reachability analysis removes the dependency on choosing an accurate model for pedestrians, however, data-driven reachability analysis can also suffer from being overly conservative~\cite{Alanwar2023}. To address the overly conservative reachable sets, authors in~\cite{Alanwar2022, Söderlund2023} explore the use of both assumed side information and behavior modes to reduce the conservativeness of the predicted reachable sets. Although there is indication that this approach can yield reachable sets that are less conservative while still providing safety benefits, a key challenge with this approach is the prediction of the behavior mode.

\begin{figure}[t]
    \centering
    \includegraphics[width=.8\linewidth]{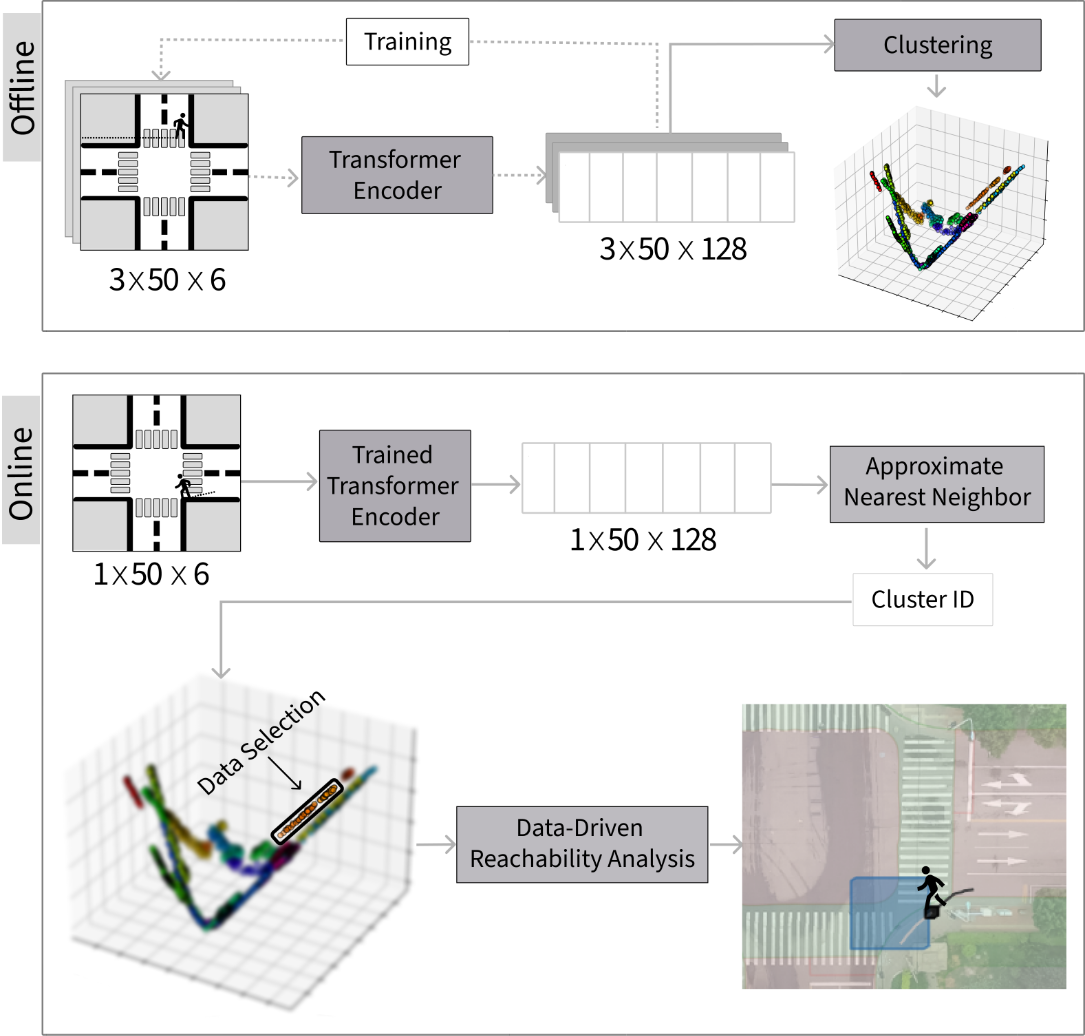}
    \caption{Overview of the proposed framework}
    \label{fig:overview}
\end{figure}

\begin{figure*}[!ht]
    \centering
    \begin{subfigure}[b]{0.448\textwidth}
        \includegraphics[width=\textwidth]{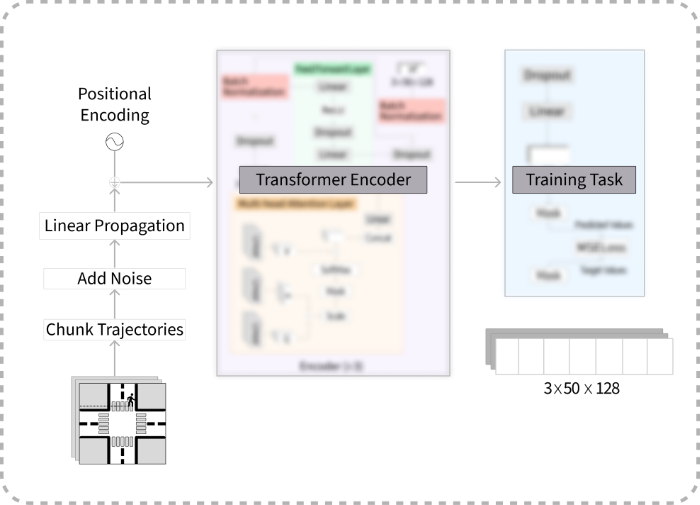}
        \caption{Pedestrian Trajectory Encoding}
        \label{fig:pipeline_a}
    \end{subfigure}
    \begin{subfigure}[b]{0.1935\textwidth}
        \includegraphics[width=\textwidth]{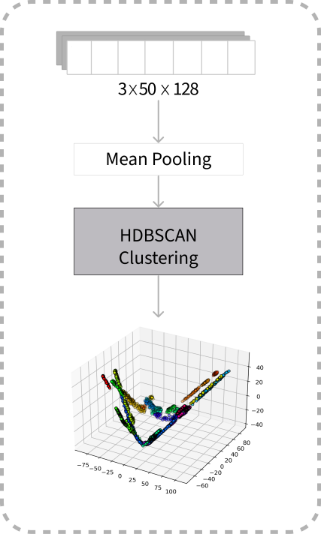}
        \caption[]{Clustering}
        \label{fig:pipeline_b}
    \end{subfigure}
    \begin{subfigure}[b]{0.342\textwidth}
        \includegraphics[width=\textwidth]{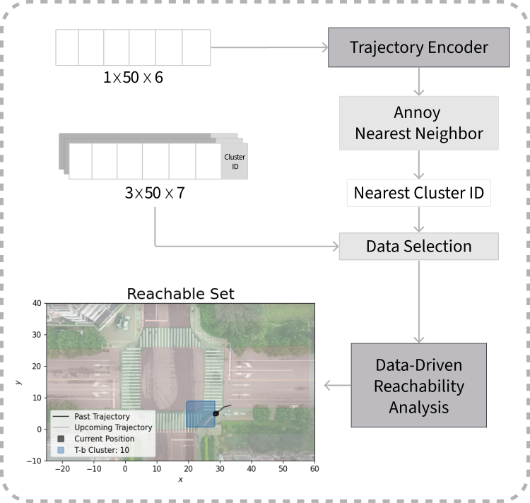}
        \caption{Data-Driven Reachability Analysis}
        \label{fig:pipeline_c}
    \end{subfigure}
    \caption{\textbf{Framework Architecture} The figure displays the three main components of the proposed framework: \textbf{(a)} Trajectory Encoding Training Process, which involves the preparation and encoding of the input trajectories to extract informative features; \textbf{(b)} the Pedestrian Behavior Clustering for grouping the encoded embeddings based on their similarity; \textbf{(c)} Data-Driven Reachability Analysis of pedestrians using data of the same cluster}
    \label{fig:pipeline}
\end{figure*}

To address this challenge, current approaches use manually engineered and labeled systems for pedestrian prediction, which often rely on fixed and pre-defined labels~\cite{Malla2020, Liang2019, Söderlund2023}. However, this manual labeling process can introduce bias and often fails to encompass the diverse range of pedestrian behaviors observed across various contexts. Alternatively, there are approaches that use clustering to identify movement patterns through various trajectory clustering techniques~\cite{Yao2017, Olive2020}. Inspired by these approaches, we are interested in exploring the use of transformers to identify clusters, since transformers have demonstrated potential in capturing complex data relationships and enhancing modeling accuracy. While there is interesting work using transformers in time series prediction~\cite{Giuliari2020, Franco2023} and improving the representation of time-series data~\cite{Zerveas2020, Lin2023}, to the extent of our knowledge, there is little previous work in the application of transformers for trajectory clustering. Thus, in this work, we explore this application and study the implications it has on data-driven reachability analysis for pedestrians.
In Figure~\ref{fig:overview}, we present an overview of our proposed approach that combines a transformer encoder with clustering and data-driven reachability analysis to predict the possible future states of pedestrians. The offline process involves training a transformer encoder to generate enriched trajectory representations, followed by clustering these encoded embeddings to create behavior clusters. Once deployed online, the system identifies the behavior cluster closest to the detected pedestrian and utilizes the data from this cluster to perform data-driven reachability analysis. By implementing this approach, we are able to utilize automatically identified behavior modes for data-driven reachability analysis.

\subsection{Contribution}
The main contribution of this paper is a pedestrian motion prediction framework that leverages transformers for clustering trajectories into behavior clusters, which are then used in data-driven reachability analysis to both reduce conservativeness and maintain safety.
Explicitly, the contributions of the paper are three-fold:
\begin{enumerate}
    \item we develop a framework that leverages transformers for clustering pedestrian trajectory data,
    \item we present an integration of the resultant behavior clusters with data-driven reachability analysis,
    \item we showcase the benefit of transformer-encoded trajectory clusters on data-driven reachability analysis using real-pedestrian trajectories.
\end{enumerate}
Additionally, the code for the evaluations performed in this work is publicly available.\footnote{\scriptsize\url{https://github.com/kfragkedaki/Pedestrian_Project}}

The remainder of the paper is organized as follows. In Section~\ref{sec:related_work}, we review the related work in trajectory clustering and transformer representation learning. In Section~\ref{sec:method}, we describe our methodology for generating enhanced trajectory representations using transformers, clustering pedestrian behaviors, and implementing data-driven reachability analysis. In Section~\ref{sec:exp}, we demonstrate our approach using real trajectory data. In Section~\ref{sec:disc}, we conclude the paper with a discussion and future directions.

\section{Related Work}\label{sec:related_work}
Due to the importance of pedestrian safety, there is an extensive body of literature exploring the use of behavior or intentions to improve the precision of pedestrian trajectory motion forecasting. For example, there are several recent proposals that incorporate pedestrian intentions to perform future trajectory predictions~\cite{Liang2019, Rasouli2019, Malla2020}. To support this work, several have developed a pre-defined set of labels and annotated datasets around pedestrian intentions and behaviors~\cite{Rasouli2017, Awad2019, Aliakbarian2019, Liu2020, Chen2021}. With a similar motivation, authors in~\cite{Söderlund2023} incorporate behavior modes into data-driven reachability analysis to further enhance the precision of the predicted future set of positions the pedestrian may occupy, assuming some level of decision uncertainty. The mentioned studies indicate the benefits of using behaviorally informed approaches for predicting the future motion of pedestrians. However, a remaining challenge with these approaches is the dependency on labels that are manually crafted based on human observations, which may not capture the full or accurate range of pedestrian behaviors and can introduce human bias.

To alleviate this challenge, we explore the use of clustering for behavior labeling. Using clustering, we are able to automatically group similar pedestrian trajectories and, potentially, reveal behavioral patterns in historical data in an unsupervised manner. When treating historical pedestrian data as time series data, there are a variety of modifications for improving the clustering quality for trajectories~\cite{ Vlachos2002, Lee2007, Sung2012, Yao2017, Wang2019, Olive2020}. Notably, authors in both~\cite{Yao2017, Olive2020} adopt auto-encoder-based machine learning approaches for automatically generating new representations of trajectory data to improve the quality of the final clusters. By incorporating computations that convert historical trajectory data into higher-dimensional feature spaces, these approaches significantly improve the quality of the clustering output in a way that avoids the need for a human designer to manually perform labeling. Inspired by these works, we also seek to explore the use of machine learning for automatically converting historical pedestrian datasets into a feature space that yields better clusters.

In particular, we explore the use of transformer models, as an alternative to the auto-encoder-based models used in~\cite{Yao2017, Olive2020}, for enhancing trajectory clustering. Transformers, initially proposed for natural language processing \cite{Vaswani2017}, have been shown to capture long-range relationships and complex patterns through their attention mechanism, effectively process sequential data, and enable parallel computation. We are specifically motivated by the recent success in utilizing transformer models for improving regression and classification of time series data~\cite{Zerveas2020}, capturing travel behavior and spatio-temporal correlations~\cite{Lin2023}, and effectively encoding transportation-related road and route representations~\cite{Chen21}. Although these applications of transformer models have shown great potential for encoding time-series and trajectory data for various tasks, the use of those models for enhancing trajectory clustering remains unexplored.

In this work, we explore an approach that leverages a transformer model to generate trajectory embeddings which can then be clustered and automatically yield behavior modes for improving pedestrian motion prediction. Specifically, we develop a framework that uses a transformer architecture similar to the one used in~\cite{Zerveas2020} to effectively cluster historical pedestrian trajectories. We consider each cluster as a ``behavior mode'' and, in the next sections, use the clusters to perform data-driven reachability analysis on pedestrians.

\section{Methodology}\label{sec:method}

The architecture of our approach is illustrated in Figure~\ref{fig:pipeline}, and consists of the pedestrian trajectory encoding process, pedestrian behavior clustering, and data-driven reachability analysis based on the identified behavior clusters. Throughout the rest of this section, we will explain each of these parts in detail.

\subsection{Pedestrian Trajectory Encoding}
\begin{figure}[!t]
    \centering
    \begin{subfigure}[b]{0.68\columnwidth}
        \includegraphics[width=\textwidth]{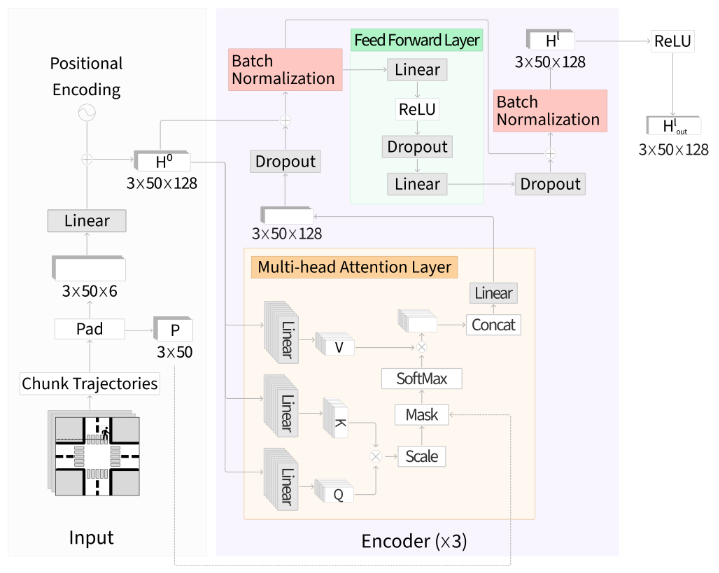}
        \caption{Transformer Encoder}
        \label{fig:input_transformer}
    \end{subfigure}
    \begin{subfigure}[b]{0.3\columnwidth}
        \includegraphics[width=\textwidth]{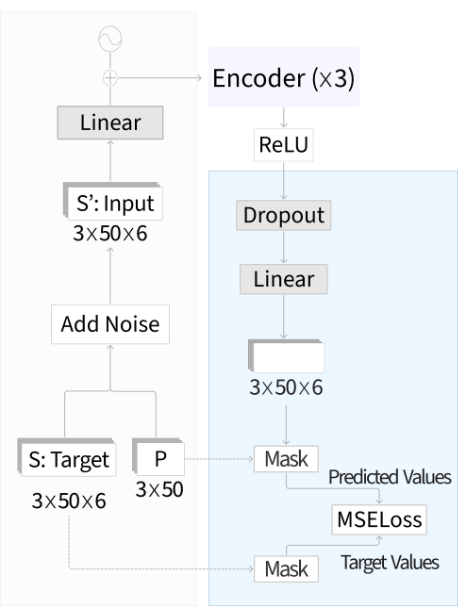}
        \caption{Training Process}
        \label{fig:transformer_training}
    \end{subfigure}
    \caption{\textbf{Trajectory Encoding Model}: The figure illustrates the encoding process of three pedestrian trajectory instances segmented to 50 time points, with six features included at each time point: \textbf{(a)} Transformer Encoder of trajectory data; \textbf{(b)} the Autoregression Training Task for unsupervised training of the transformer encoder.}
    \label{fig:trajectory_encoding}
\end{figure}

The first part of the system is the pedestrian trajectory encoding which refers to the unsupervised training of the trajectory encoder. In Figure~\ref{fig:trajectory_encoding}, the transformer encoder and its training process are shown for three trajectories, and the details of the trajectory input preparation, the encoder model, and the training process are further explained below.

\subsubsection{Trajectory Input Preparation}

For this work, we utilize trajectory data consisting of the pedestrians' positions, velocities, and accelerations in both spatial dimensions, forming a 6-dimensional feature vector at each time point in the sequence. We develop our approach around these data points since this is the most commonly detected motion data by pedestrian detection systems. To address the variability in behaviors observed over longer trajectories, the data is segmented into smaller chunks of specified length $d_c$. We standardize the length of all input sequences by padding shorter trajectories with zeros, creating a binary matrix $P$ to distinguish actual data from padded values in the model. Finally, each feature vector is projected from a 6-dimensional to a 128-dimensional space with positional encodings added to retain the sequential nature of the data, addressing the transformer architecture's inherent lack of input order consideration. The input to the transformer model is then the high-dimensional sequence matrix $H^{0} \in \mathbb{R}^{d_b \times d_c \times 128}$, where $d_b$ refers to the batch number of trajectory instances being processed and $d_c$ to the chunk size of the trajectories.

\subsubsection{Transformer Encoder}
A transformer is a type of deep neural network that uses attention mechanisms to capture dependencies between elements in a sequence. In natural language processing (NLP), transformers are used to understand the context of a sentence, where the meaning of a word depends on its relationship with other words. For instance, the sentences ``Where is the bus stop?'' and ``How does the bus stop?'', despite containing similar words, have different meanings due to the order and relationships of the words. In our context, we implement the transformer architecture as described by~\cite{Vaswani2017, Zerveas2020} to process pedestrian trajectory data. Similar to how words in a sentence have contextual relationships, each time point in a pedestrian's trajectory is linked to preceding and succeeding movements, and by comparing them, features like the distance covered, the speed changes, or the walking style of the pedestrian can be inferred.

To perform this inference for pedestrian trajectories, we use the transformer encoder in Figure~\ref{fig:input_transformer}. As displayed in the figure, the higher dimensional trajectory embeddings $H^{0}$ are fed into a stack of three transformer encoder layers. Each layer comprises a pair of sublayers: a multi-head attention (MHA) and a fully connected feed-forward (FF) sublayer. We apply dropout~\cite{srivastava2014dropout} to both sublayers and then incorporate a skip connection~\cite{Residuals2015} by adding the input of the sublayer to its output, followed by batch normalization (BN)~\cite{Ioffe2015}. The MHA sublayer utilizes a self-attention network with eight heads, enabling the model to capture different trajectory-related information simultaneously. On the other hand, the FF sublayer processes each position in the trajectory independently, applying the same weights across all positions to learn position-related information.

By using a stack of transformer encoder layers, the output embeddings from one layer serve as the input to the next layer, allowing the model to build upon the information learned in the previous layers. For example, if the model initially learns to identify basic characteristics such as distance covered and speed changes in early layers, then in the subsequent layers, it could use this foundational knowledge to infer more complex behaviors, such as determining whether or not a pedestrian is in a rush. This progressive learning and integration of information across layers enable a nuanced understanding of pedestrian dynamics that can be difficult for single-layer models to achieve. The high-dimensional embeddings extracted from the last layer of the encoder with ReLU activation applied, $H^l_{out}$, are the encoded trajectory embeddings used to identify behavior clusters in the next phases. 

\subsubsection{Autoregression Training Task}

For training the encoder, we utilize an unsupervised learning approach based on the autoregression denoising task proposed by~\cite{Zerveas2020}. As shown in Figure~\ref{fig:transformer_training}, we add noise to the input data, by setting portions of it to zero, and task the model with predicting these masked values. The noised input is fed into the encoder, and the output embeddings of the trajectory encoder with dropout applied,  $H^l_{out}$, are then mapped back to the 6-dimensional feature space. These outputs are compared with the original non-noised values to evaluate the model's predictions. We compute the Mean Squared Error (MSE) Loss, which measures the accuracy of predictions across all non-padded values, in contrast to ~\cite{Zerveas2020} who focus solely on the masked data.
This task aims for the encoder to learn from unlabeled data to generate informative trajectory embeddings.

\subsection{Pedestrian Behavior Clustering}

For clustering the encoded trajectory embeddings, we have chosen Hierarchical Density-Based Spatial Clustering of Applications with Noise (HDBSCAN)~\cite{HDBSCAN} due to its robustness in identifying clusters of arbitrary shapes and sizes. HDBSCAN does not require predefining the number of clusters, unlike other clustering algorithms such as K-means. Instead, it works by assessing the density of data points within a space and dynamically adjusting clusters based on a density threshold $min\_sample$.

In the clustering process, the high-dimensional embeddings $H^l_{out}$ extracted from the trajectory encoding phase are used. We apply mean pooling across the time dimension of these embeddings to aggregate the temporal information into a single representative vector for each trajectory. These pooled embeddings are then fed into the HDBSCAN algorithm to identify and group similar patterns of pedestrian behavior movements together. 

\subsection{Data-Driven Reachability Analysis}

We perform data-driven reachability analysis of pedestrians using the clusters identified from the encoded trajectory embeddings. Our approach builds upon the work of~\cite{Söderlund2023}, who used historically similar trajectories to predict future states. However, we enhance their framework by using automatically-identified behavior clusters instead of manually crafted labels. To predict the reachable set of a pedestrian, we first encode its trajectory using our trained transformer encoder. Then, we employ an Approximate Nearest Neighbor (ANN) algorithm to identify to which cluster the unseen trajectory most closely belongs. For brevity, we do not include the full formulation of the data-driven reachability analysis. For readers who would like to see the details of the reachability analysis, we direct them to~\cite[Algorithm 1]{Söderlund2023}, where the input $\mathcal C$ is the cluster of trajectories identified by ANN.

In this paper, we present the evaluation of only inputting the nearest cluster identified by ANN into the data-driven reachability analysis. Importantly, we note that the data-driven reachability analysis also supports the input of multiple clusters appended together. By increasing the number of clusters given to the data-driven reachability analysis, we increase both the conservativeness and safety of the predicted reachable sets. At the extreme, the inclusion of all historical data yields results similar to~\cite{Alanwar2023}, which computes a reachable set that accounts for all possible linear models that are consistent with the historical data. Since the design of how many of the behavior clusters are selected is dependent on the particular application, we only evaluate the results of picking the closest one. An important future direction will be to develop schemes for adapting the number of chosen clusters based on the scenario.

\section{Evaluation}\label{sec:exp}

In this section, we present the trajectory encoding training and clustering results, and perform data-driven reachability analysis using the identified clusters. We also describe the dataset and experimental setup that were employed to conduct the evaluations presented.

\subsection{Pedestrian Trajectory Dataset}

We use the SIND dataset~\cite{SinD} to train and evaluate the transformer encoder and clustering on the work of~\cite{Söderlund2023}. The SIND dataset is an open-source signalized dataset of real road users' trajectories from a large intersection in Tianjin, China, captured by a drone. We specifically focus on the pedestrian trajectories, which provide information about the spatial location, the velocity, and the acceleration over each time point in the trajectory. The location $(x, y)$ is bounded by the spatial map limit of the dataset, and the velocity $(v_x, v_y)$ and the acceleration $(a_x, a_y)$ in both spatial dimensions are included. The trajectory data are filtered for falsely recognized pedestrians by completely removing stationary detections where all values of $v_x$ and $v_y$ in the trajectory equal zero.

As discussed in Section~\ref{sec:method}, the data is segmented into chunks with a trajectory length of $d_c=50$, and padded to standardize shorter trajectories. This differs from the work of~\cite{Söderlund2023} who use a chunk size of 90 and split the trajectories utilizing a sliding window technique of size one. By shortening the length of chunks and avoiding the sliding window approach, we aim to capture distinct pedestrian behaviors and reduce biases in the data.
For training, we introduce noise to the input data following the mechanism introduced by~\cite{Zerveas2020} that randomly masks a proportion $r$ of trajectory data based on a Bernoulli distribution with mean length $l_m$. To increase the complexity of the prediction task, both the mean length $l_m$ of sequential input data masked and the proportion $r$ of the data masked are incrementally increased during the training process.

The dataset is split into training, validation, and testing subsets, comprising 70\%, 20\% and 10\% of the chunked data, respectively. The training and validation subsets are randomly split and used for training the transformer encoder. In the clustering phase, both the training and validation subsets are considered historical data, and are used to identify patterns and create the behavior clusters. The testing subset, which is fixed, is used exclusively for evaluating the effectiveness of the behavior clusters in the data-driven reachability analysis.

\subsection{Experimental Setup}

To evaluate the effectiveness of the proposed framework, we compare the following four approaches for selecting historical data to perform data-driven reachability analysis for a detected pedestrian.

\begin{enumerate}
    \item \textbf{Baseline Method~\cite{Söderlund2023}}: We utilize all historical data that intersect with the detected pedestrian's position.
    \item \textbf{Labeling Method~\cite{Söderlund2023}}:  We split historical data into six predefined behavior modes, and the data of the mode of the detected pedestrian is utilized. We also apply a location and heading filter to the selected data.
    \item \textbf{Non-encoded Trajectory Clustering}: We cluster historical data using the original 6-dimensional features and utilize all data in the detected pedestrian's cluster.
    \item \textbf{Transformer-Encoded Trajectory Clustering}: We cluster historical data using encoded trajectory embeddings generated by the trained encoder of Figure~\ref{fig:input_transformer}. We utilize all data in the detected pedestrian's cluster.
\end{enumerate}

To ensure a fair comparison between these fundamentally different methods, we apply a cluster distance filter in both clustering techniques. Moreover, if the data provided for performing reachability analysis in any of the four methods exceeds the memory limitations of the machine running the analysis or if there is no data satisfying the above criteria, that trial is completely excluded. This is practically motivated, since in cases where memory is exceeded or no data is available, a fallback reachability analysis would be performed instead.

\begin{figure}[!t]
    \centering
    \includegraphics[width=.52\linewidth]{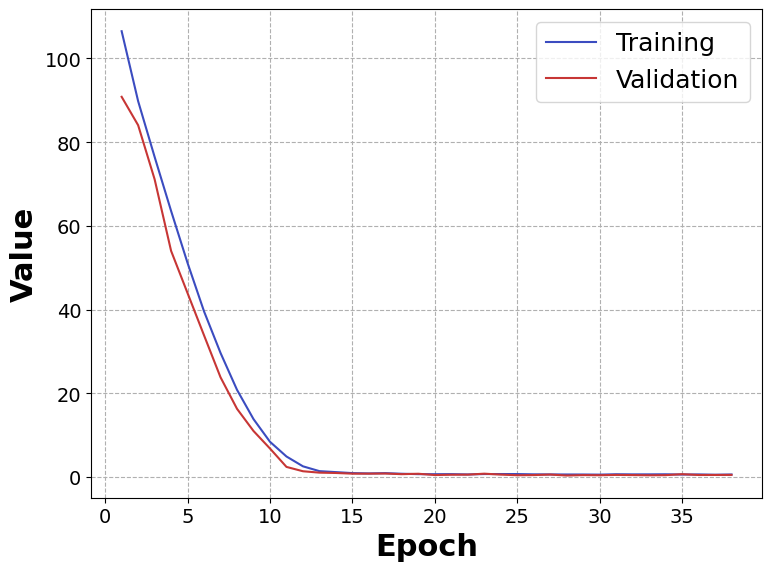}
    \caption{Transformer encoder training and validation loss.}
    \label{fig:training}
\end{figure}

The implementation of our code is publicly available\footnote{\scriptsize\url{https://github.com/kfragkedaki/Pedestrian_Project}}. For the transformer-encoded trajectory clustering approach, we implemented our transformer encoder using PyTorch,
and performed hyperparameter tuning with Ray, selecting parameters based on the MSELoss on the validation subset. Hyperparameter tuning was conducted on an Ubuntu system equipped with an Intel Core i7-8850H CPU at 2.60GHz, 12 GB of memory, and an NVIDIA GeForce RTX 2080 Mobile GPU. The chosen model was trained for 37 epochs using the Adam optimizer with a batch size of 256, a learning rate of \(5.011 \times 10^{-4}\), L2 regularization of 0.05, a dropout rate of 0.1, and a ReLU activation function. To evaluate the learning curve of our model, we monitored MSE Loss on both the training and validation subsets. As shown in Figure~\ref{fig:training}, the MSE Loss decreases over time, indicating that the model is effectively learning to generalize to unseen data. For both clustering methods, we utilized the HDBSCAN algorithm from scikit-learn,
while for the nearest neighbor cluster searches the Annoy\footnote{\scriptsize\url{https://github.com/spotify/annoy}} library.

\subsection{Results}

In this subsection, we present the results of our approach, showing the clusters created by both clustering methods and applying data-driven reachability analysis using all four aforementioned approaches. 

\begin{figure*}[!ht]
    \centering
    \begin{subfigure}[b]{0.32\textwidth}
        \includegraphics[width=\textwidth]{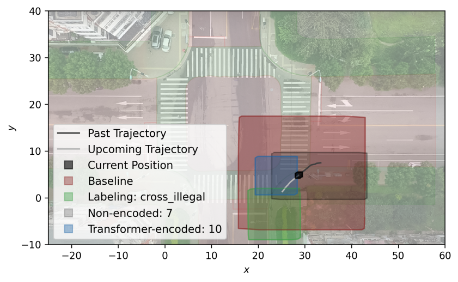}
        \caption{Cross Illegal}
        \label{fig:cross_illegal}
    \end{subfigure}
    \begin{subfigure}[b]{0.32\textwidth}
        \includegraphics[width=\textwidth]{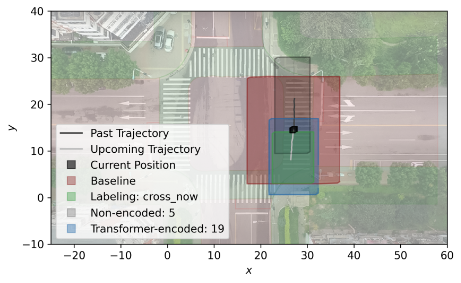}
        \caption{Cross Now}
        \label{fig:cross_now}
    \end{subfigure}
    \begin{subfigure}[b]{0.32\textwidth}
        \includegraphics[width=\textwidth]{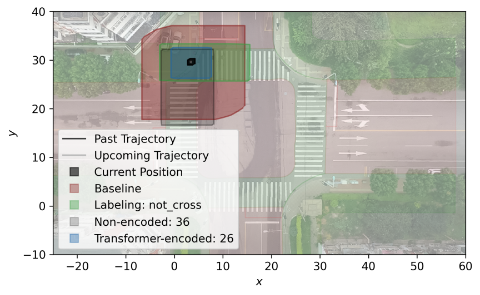}
        \caption{Not Cross}
        \label{fig:not_cross}
    \end{subfigure}
    \caption{Reachable sets for three different scenarios using historical trajectories based on: the baseline and labeling oracle defined by~\cite{Söderlund2023}, a non-encoded trajectory clustering, and a transformer-encoded trajectory clustering approach.}
    \label{fig:reachability_analysis}
\end{figure*}

\subsubsection{Trajectory Clusters}

\begin{figure}[!b]
    \centering
    \subfloat[Initial Data]{
        \includegraphics[width=0.40\linewidth]{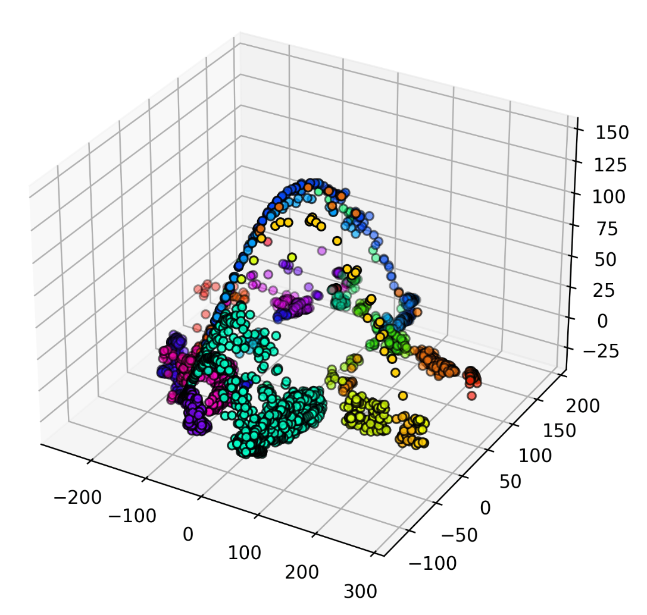}
        \label{fig:clustering_original}
    }
    \subfloat[Encoded Data]{
        \includegraphics[width=0.40\linewidth]{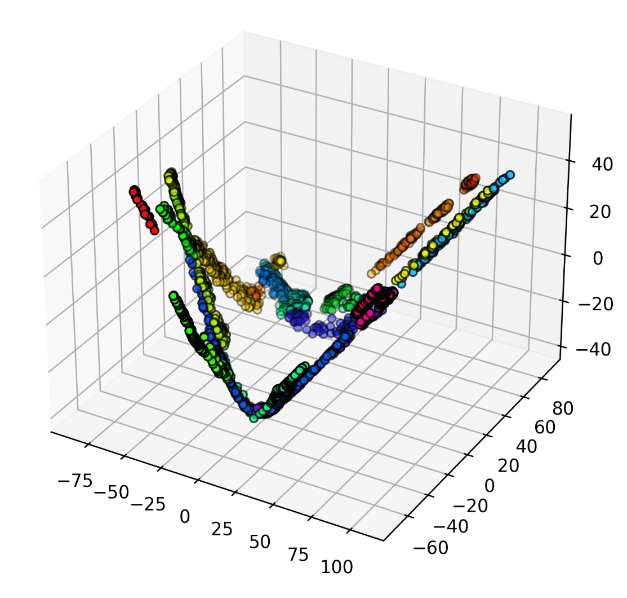}
        \label{fig:clustering_encoded}
    }
    \caption{PCA on the initial data and the encoded trajectory embeddings, color-coded by cluster.}
    \label{fig:clustering_results}
\end{figure}

After clustering both the initial trajectory data and the encoded trajectory embeddings, we applied Principal Component Analysis (PCA)
to reduce the dimensionality for visualization and analysis. The initial trajectory data, consisting of position, velocity, and acceleration, was reduced from 6 dimensions to 3 dimensions, while the encoded trajectory embeddings were reduced from 128 dimensions to 3 dimensions. Figure~\ref{fig:clustering_results} displays both PCA-transformed data points, color-coded according to their respective clusters. We note that the PCA of the non-encoded initial data corresponds fairly closely to the geometry of the intersection, indicating a strong reliance on the position of the pedestrian for predicting its behavior. In contrast, the PCA of the encoded data shows a different structure, mapped to a narrower range of values. These visualizations emphasize that the encoded data differs from the initial trajectory data in structure and is picking up patterns in the data that are possibly more rich than just the position of the pedestrian.

\subsection{Data-Driven Reachability Analysis Evaluation}

We examine the reachability of pedestrian states using the identified transformer-encoded trajectory clusters and compare it with the other three methods.
Figure~\ref{fig:reachability_analysis} visualizes the reachable sets for three scenarios: (a) the pedestrian is crossing illegally (labeled ``cross illegal''), (b) the pedestrian is crossing legally (labeled ``cross now''), and (c) the pedestrian is not currently crossing (labeled ``not cross''). In Figure~\ref{fig:cross_illegal}, we see that the reachable set computed from the transformer-encoded trajectory cluster is clearly the most accurate and least conservative.
In Figure~\ref{fig:cross_now}, we see that the reachable set computed from the manually labeled data is the most accurate and least conservative, while the reachable set computed from the transformer-encoded trajectory cluster is comparable. This particular scenario provides a preliminary indication that, although a human can spend time manually creating clusters that may perform better, similar performance can be achieved in a fully automated fashion.
In Figure~\ref{fig:not_cross} in which the pedestrian does not cross, we see that the transformer-encoded trajectory clustering approach is able to predict this behavior and generate an intuitive reachable set. Qualitatively, we can see that in these scenarios, the reachable sets generated from the transformer-encoded trajectory clusters could result in more efficient traffic flow, since vehicles can more precisely plan around the future motion of pedestrians.
Quantitatively, based on the volume of each method in the scenarios and the average volumes on the testing data presented in Table~\ref{table:zonotope_areas}, we can conclude that the baseline method is clearly overly conservative, while the transformer-encoded trajectory clustering method is competitive with the labeling approach, while outperforming the non-encoded trajectory clustering.

\begin{table}[!t]
\renewcommand{\arraystretch}{1.3}
\caption{Comparison of zonotope area volumes in the scenarios, and the average zonotope volumes on the testing dataset.}
\label{table:zonotope_areas}
\centering
\resizebox{\columnwidth}{!}{%

\begin{tabular}{@{}l|cccc@{}}
\toprule
 \multicolumn{1}{c}{} & \textbf{Baseline (m\textsuperscript{2})} & \textbf{Labeling (m\textsuperscript{2})} & \textbf{\makecell{Non-encoded (m\textsuperscript{2})}}& \textbf{\makecell{Transformer-encoded (m\textsuperscript{2})}}\\
\hline
Cross Illegal & 657.0511 & 122.0815 & 204.1029 & \textbf{73.8813} \\
Cross Now & 454.6648 & \textbf{98.4315} & - & 172.4226 \\
Not Cross & 396.1144 &  141.6194 & 169.2804 & \textbf{52.6826} \\
\bottomrule
\makecell{Average Volume} & 309.2136 &  175.6101 & 320.1112 & 259.526 \\

\bottomrule
\end{tabular}

}
\end{table}

\begin{figure}[!t]
    \centering
    \includegraphics[width=.8\linewidth]{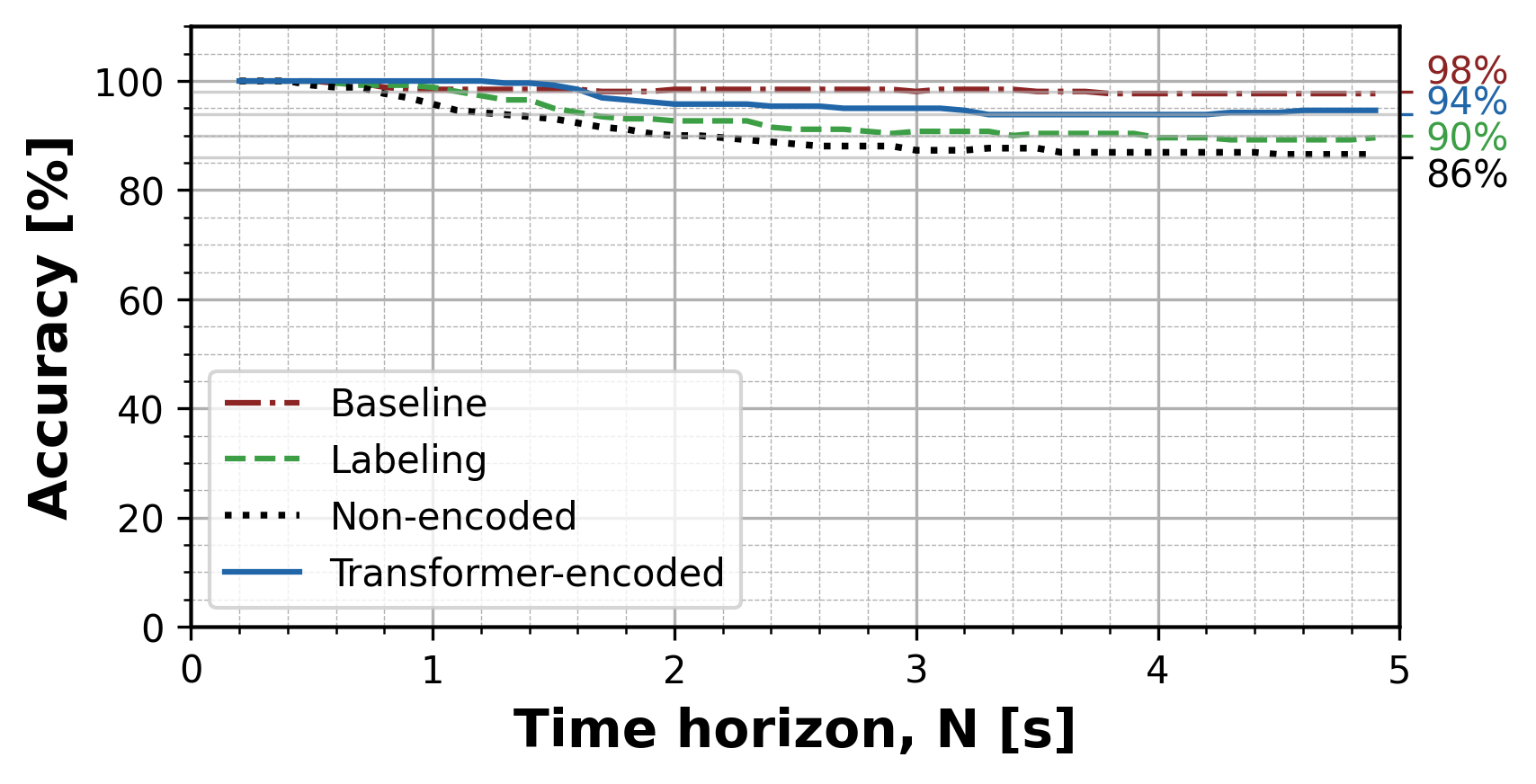}
    \caption{State inclusion accuracy on the testing dataset.}
    \label{fig:accuracy_reachbility_analysis}
\end{figure}

Additionally, to assess the overall safety of each method, we evaluate each method's state inclusion accuracy, shown in Figure~\ref{fig:accuracy_reachbility_analysis}. This metric represents how frequently the actual future state of a pedestrian falls within the predicted reachable set.
The transformer-encoded trajectory clustering approach achieves the best performance after the overly conservative baseline, while the non-encoded trajectory clustering performs the worst out of all four methods. For evaluating these results, impractical outlier cases needed to be fully removed from the experiments. For brevity, we do not report the statistics of these outlier cases and direct interested readers to the open-source implementation for more details.

\section{Discussion \& Future Work}\label{sec:disc}




In this work, we propose a transformer-encoded trajectory clustering approach that automatically selects historical trajectories for data-driven reachability analysis of pedestrians. This approach is shown to be effective in maintaining safety while enhancing the precision of pedestrian motion predictions. The transformer-based method offers a clear advantage over classical clustering by capturing complex data relationships and improving modeling accuracy. From our findings, we observe that the reachability analysis can face challenges due to either insufficient or excessive historical data, and there is more work to be done on how to safely handle these cases.
Finally, since the SIND dataset captures pedestrian motion from the perspective of a drone, evaluating our approach on data captured from a vehicle's perspective is important to understand its applicability in automated vehicles. This evaluation will help determine the feasibility and effectiveness of our transformer-encoded trajectory clustering in real-world scenarios involving autonomous driving.

\balance


\bibliographystyle{IEEEtran}
\bibliography{references}




\end{document}